\documentclass[10pt,twocolumn,letterpaper]{article}

\usepackage[pagenumbers]{cvpr} 

\usepackage{fancyhdr}

\usepackage[dvipsnames]{xcolor}

\usepackage{units}
\usepackage{amsmath}
\usepackage{multirow}
\usepackage{arydshln}
\usepackage{booktabs}
\usepackage{graphicx}
\usepackage{comment}
\usepackage[accsupp]{axessibility} 

\definecolor{cvprblue}{rgb}{0.21,0.49,0.74}
\usepackage[pagebackref,breaklinks,colorlinks,citecolor=cvprblue]{hyperref}

\usepackage{titling}

\fancypagestyle{firstpage}{
  \fancyhf{} 
  \fancyhead[C]{To appear in Proceedings of the 2024 IEEE/CVF Conference on Computer Vision and Pattern Recognition (CVPR). \\ The conference version will be available soon.} 
}

\newcommand{\new}[1]{{#1}} 

\def\paperID{8591} 
\def\confName{CVPR}
\def\confYear{2024}

\title{XFeat: Accelerated Features for Lightweight Image Matching\vspace*{-0.6cm}}

\date{}

\author{
Guilherme Potje$^1$ \hspace{0.3in}
Felipe Cadar$^{1,2}$    \hspace{0.3in}
André Araujo$^3$    \vspace{0.02in}\\ 
Renato Martins$^{2,4}$   \hspace{0.3in}
Erickson R. Nascimento$^{1,5}$ \vspace{0.1in} \\
$^1$Universidade Federal de Minas Gerais \hspace{0.06in}
$^2$Université de Bourgogne, ICB UMR 6303 CNRS\\
$^3$Google Research 
 \hspace{0.08in} $^4$Université de Lorraine, LORIA, Inria  \hspace{0.08in} $^5$Microsoft \\
{\tt\small \{guipotje,cadar,erickson\}@dcc.ufmg.br, renato.martins@u-bourgogne.fr, andrearaujo@google.com}\vspace*{-0.5cm}
}

\begin{document}
\maketitle


\begin{abstract}



We introduce a lightweight and accurate architecture for resource-efficient visual correspondence. Our method, dubbed XFeat (Accelerated Features), revisits fundamental design choices in convolutional neural networks for detecting, extracting, and matching local features. Our new model satisfies a critical need for fast and robust algorithms suitable to resource-limited devices. In particular, accurate image matching requires sufficiently large image resolutions -- for this reason, we keep the resolution as large as possible while limiting the number of channels in the network. Besides, our model is designed to offer the choice of matching at the sparse or semi-dense levels, each of which may be more suitable for different downstream applications, such as visual navigation and augmented reality. Our model is the first to offer semi-dense matching efficiently, leveraging a novel match refinement module that relies on coarse local descriptors. XFeat is versatile and hardware-independent, surpassing current deep learning-based local features in speed (up to 5x faster) with comparable or better accuracy, proven in pose estimation and visual localization. We showcase it running in real-time on an inexpensive laptop CPU without specialized hardware optimizations. Code and weights are available at \url{www.verlab.dcc.ufmg.br/descriptors/xfeat_cvpr24}.

\end{abstract}
\thispagestyle{firstpage} 
\section{Introduction}
\label{sec:intro}

\begin{figure}[t]
	\centering
	\includegraphics[width=0.99\columnwidth]{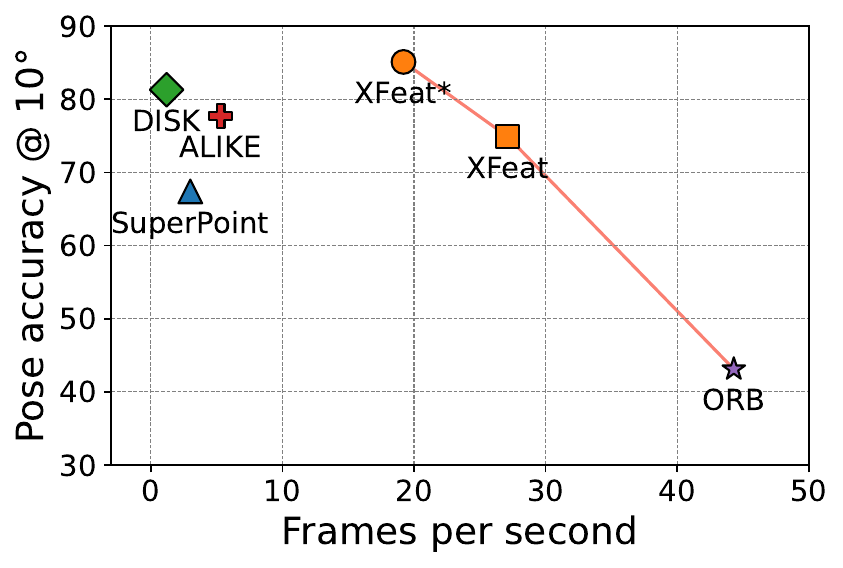}
	\caption{{\bf{In XFeat, accuracy meets efficiency.}} XFeat delivers great trade-off between speed and relative pose estimation accuracy on the Megadepth-1500 dataset, as evidenced by the Pareto-frontier curve in orange. Its lightweight architecture enables real-time feature extraction on GPU-free settings and resource-constrained devices without hardware-specific optimizations. Inference speed on a budget-friendly laptop (\texttt{Intel(R) i5-1135G7 @ 2.40GHz CPU}) at VGA resolution. $^*$ denotes semi-dense extraction.} \vspace*{-0.2cm} 
	
	\label{fig:teaser}
\end{figure}

As a crucial step for many higher-level vision tasks, local image feature extraction remains a highly active topic of research. Despite the recent advancements, the large improvements achieved from recent image matching methods~\cite{cit:dkm, cit:loftr, cit:aspanformer, cit:pdcnet} mostly come at the cost of high computational requirements and increased implementation complexity. 
Since image feature extraction is critical for a myriad of tasks~\cite{cit:hloc, cit:orbslam, cit:visualsfm, cit:potje_sfm, cit:colmap, cit:cann, cit:delf}, efficient solutions are highly desirable, especially on resource-constrained platforms such as mobile robots, augmented reality, and portable devices, where scarce computational resources are often allocated to multiple tasks simultaneously. Although specific works aim to perform hardware-level optimization for existing architectures~\cite{cit:zippypoint}, which is still hardware-specific and cumbersome in practice, few works focus on the architectural design for efficient feature extraction~\cite{cit:alike}.


Drawing inspiration from the state-of-the-art developments in several fronts of image matching, we present \textbf{XFeat}: a novel convolutional neural network (CNN) architecture that performs keypoint detection and local feature extraction using carefully designed strategies to reduce computational footprint as much as possible, while being robust and accurate. 
XFeat is designed to be hardware-agnostic, ensuring broad applicability across platforms, but this does not preclude the potential for optimizing XFeat on specific hardware configurations. 
Moreover, XFeat is suitable to perform both sparse feature matching based on keypoints and dense matching of the coarse feature map. This versatility brings the best of both worlds: keypoint-based methods are more suitable to efficient visual localization based on Structure-from-Motion (SfM) maps~\cite{cit:hloc}, while dense feature matching can be more effective for relative camera pose estimation in poorly textured scenes~\cite{cit:loftr, cit:aspanformer}. 


\begin{figure}[t]
	\centering
	\includegraphics[width=0.99\columnwidth]{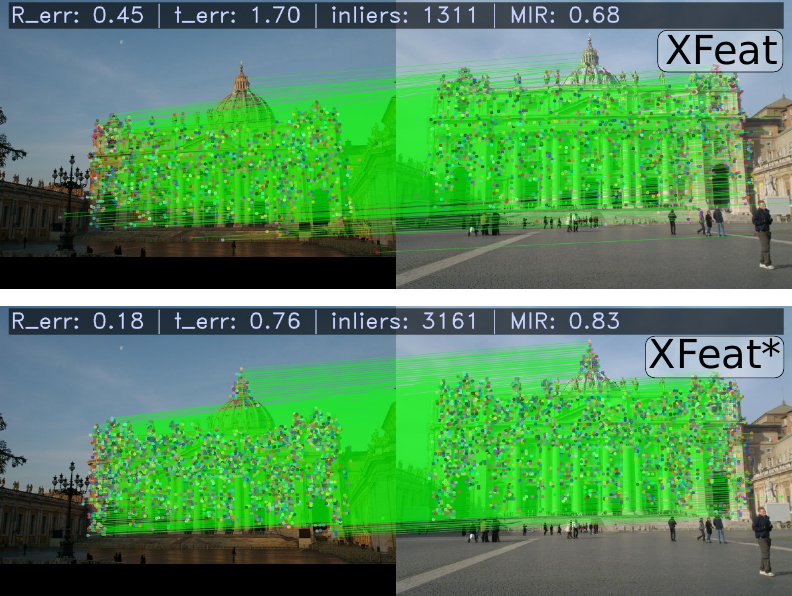}
	\caption{{\bf{Sparse (top) and semi-dense (bottom) matching.}} XFeat stands out with its dual ability to perform both sparse and semi-dense matching, providing fast features for a wide range of applications from visual localization with sparse matches to pose estimation and 3D reconstruction where denser correspondences deliver additional constraints and a more complete representation.} \vspace*{-0.2cm} 
	\label{fig:teaser_match}
\end{figure}

Compared with current methods available for image correspondence, our method significantly improves the trade-off ratio between matching accuracy and computational efficiency as shown in~\cref{fig:teaser}, outperforming all lightweight deep learning local feature alternatives by up to $5\times$ in speed while being comparable to much larger models as SuperPoint~\cite{cit:superpoint} and DISK~\cite{cit:disk} in accuracy. To mitigate computational costs while maintaining competitive accuracy, our work brings three main contributions:
\begin{itemize}
    \item A novel lightweight CNN architecture that can be deployed on resource-constrained platforms and downstream tasks that require high throughput or computational efficiency, without the requirement of time-consuming hardware-specific optimizations. Our method can readily replace existing lightweight handcrafted solutions~\cite{cit:orb}, expensive deep models~\cite{cit:disk, cit:superpoint} and lightweight deep models~\cite{cit:alike} in several downstream tasks such as visual localization and camera pose estimation; 
    \item We design a minimalist, learnable keypoint detection branch that is fast and suitable for small extractor backbones, showing its effectiveness in visual localization, camera pose estimation, and homography registration;
    \item Lastly, a novel match refinement module for obtaining pixel-level offsets from coarse semi-dense matches is proposed. Our new strategy does not require high resolution features besides the local descriptors themselves as opposed to existing techniques~\cite{cit:loftr, cit:aspanformer}, greatly reducing compute and achieving high accuracy and matching density shown in \cref{fig:teaser}, and \cref{fig:teaser_match} respectively. 
\end{itemize}



\section{Related Work}

\paragraph{Image matching.}
Modern image matching techniques range from employing classic keypoint detection~\cite{cit:sift, cit:harris} coupled with deep-learning based description of local patches~\cite{cit:hardnet, cit:affnet, cit:geobit, cit:geopatch, cit:deal}, to performing joint keypoint detection and description~\cite{cit:superpoint, cit:r2d2, cit:disk, cit:dalf} in the same CNN backbone. More recently, middle-end approaches, known as learned matchers~\cite{cit:superglue, cit:lightglue, cit:lfm3d}, and also end-to-end semi-dense~\cite{cit:loftr, cit:aspanformer} and dense~\cite{cit:pdcnet, cit:dkm} methods, demonstrated remarkable improvements in robustness and accuracy for matching wide-baseline image pairs, especially with the recent advances introduced by the transformer architecture~\cite{cit:transformer}. However, recent methods largely emphasize image matching accuracy and robustness, thereby inflating computational demands to undesired levels, even for systems with moderate GPU resources. They require significant adaptations to work efficiently in large-scale downstream tasks such as visual localization \cite{cit:hloc}, simultaneous localization \& mapping~\cite{cit:orbslam}, and structure-from-motion~\cite{cit:colmap}. In contrast, in this paper we show that it is possible to drastically reduce compute utilization in both sparse keypoint extraction and pixel-level semi-dense matching, while attaining similar, or even better performances compared to more computationally expensive methods.

\vspace{-0.3cm}
\paragraph{Efficient description \& matching.}
 Recent works highlight the growing emphasis on computational efficiency for description and matching. SuperPoint~\cite{cit:superpoint} proposed a self-supervised CNN for both keypoint detection and description. However, one major disadvantage of using SuperPoint is that it can still incur significant computational costs when applied to image sizes that are common for image matching. 
 SiLK~\cite{cit:silk} reevaluates elements of learned feature extraction, proposing an effective yet simple strategy for keypoint and descriptor learning that achieves performance comparable to existing methods.
 The key aspect that underscores SiLK’s competitiveness -- its dependence on the original image size for descriptor extraction -- is also its main drawback in terms of computational cost, as it substantially slows down inference.
 ALIKE~\cite{cit:alike} introduced a lightweight network balancing robustness and speed, with differentiable keypoint detection and a neural reprojection loss. Yet, its reliance on the original image resolution in the final feature map considerably increases memory and compute footprints.
 ZippyPoint~\cite{cit:zippypoint} incorporates quantization and binarization in a CNN. Although it achieved notable speed improvements, it requires custom compilation and specific low-level processor arithmetic operations, restricting its applicability across diverse hardware.
 
 Works considering minimalist CNN architectures may employ both fixed handcrafted and learned filters in convolutional blocks~\cite{cit:keynet}. Beyond feature extraction, recent advancements in feature matching also highlight the necessity for quick inference speeds. LightGlue~\cite{cit:lightglue} speeds up learnable feature matching and maintains high accuracy compared to SuperGlue~\cite{cit:superglue}. 
 Nevertheless, LightGlue's transformer-based architecture is still costly for tasks where computational efficiency is critical.
 In contrast to existing methods, we focus on highly-efficient and robust image matching for ubiquitous deployment: from resource-limited devices such as low-budget boards and embedded systems to smartphones and cloud applications.

\section{XFeat: Accelerated Features}

\begin{figure*}[t]
	\centering
	\includegraphics[width=0.9\textwidth]{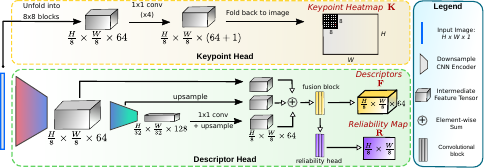}

	\caption{{\bf {Accelerated feature extraction network architecture.}} 
	XFeat extracts a keypoint heatmap $\mathbf{K}$, a compact 64-D dense descriptor map $\mathbf{F}$, and a reliability heatmap $\mathbf{R}$. It achieves unparalleled speed via early downsampling and shallow convolutions, followed by deeper convolutions in later encoders for robustness. Contrary to typical methods, it separates keypoint detection into a distinct branch, using $1 \times 1$ convolutions on an $8 \times 8$ tensor-block-transformed image for fast processing.
    }
	\label{fig:method}
\vspace{-0.2cm}
\end{figure*}

Local feature extraction accuracy heavily depends on input image resolution. For instance, in camera pose, visual localization, and SfM tasks, the correspondences should be fine-grained enough to allow pixel-level matches. However, feeding high-resolution images into network backbones increases computational requirements to undesired levels even for simple, small network backbones such as SuperPoint VGG-like architecture~\cite{cit:vgg, cit:superpoint}. In this section, we describe how to reduce significantly the computational cost using strategies to minimize the computational budget while mitigating robustness loss due to a considerably smaller CNN backbone.

\subsection{Featherweight Network Backbone}
Let $\mathbf{I} \in \mathbb{R}^{H \times W \times C}$ be a gray-scale image, where $H$ is the height, $W$ the width in pixels, and $C=1$ denotes the number of channels.
To decrease a CNN processing cost, a common approach is to start with shallow convolutions and then incrementally halve spatial dimensions $(H_i,W_i)$ while doubling the channel count $C_i$ in the $i$-th convolutional block~\cite{cit:vgg}. Assuming a convolutional layer with unit stride, padding, no bias term and square kernel size $k \times k$, the cost of convolution in terms of floating point operations ($F_{ops}$) for the $i$-th layer can be expressed as:
\begin{equation}
    F_{ops} = H_i \cdot W_i \cdot C_i \cdot C_{i+1} \cdot k^2.
    \label{eqn:conv}
\end{equation}
Naively pruning channels $C$ across the entire network compromises its capability of handling challenges like varying illumination and viewpoint as demonstrated in the ablation experiments (\cref{sec:ablations}).

Efficient networks~\cite{cit:mobilenet, cit:shufflenet} use depthwise separable convolutions to cut down $F_{ops}$ by up to 9 times (with $3 \times 3$ kernel size) with fewer parameters than standard convolutions. However, in local feature extraction, where shallower networks handle larger image resolutions~\cite{cit:superpoint, cit:r2d2, cit:silk, cit:alike, cit:aslfeat}, this approach is less effective compared to their original use in low-resolution input scenarios like classification and object detection~\cite{cit:vgg, cit:mobilenet, cit:resnet}. This leads to limited representational capacity and minor speed gains in shallow networks for local feature extraction.

In \cref{eqn:conv}, the $H_i * W_i$ terms emerge as the primary computational bottleneck impacting $F_{ops}$ in CNNs. SuperPoint~\cite{cit:superpoint} and ALIKE~\cite{cit:alike} reduce channel depth and layer count uniformly to alleviate the problem. We delve into the core of the issue, formulating  a strategy to minimize early-layer depth and reconfigure channel distribution, significantly improving the accuracy-compute trade-off. Our proposed strategy involves reducing the channel count in initial convolution layers as much as possible due to the high spatial resolution. To counterbalance the parameter reduction, rather than adhering to the traditional VGG-like approach~\cite{cit:vgg} of doubling channels, we propose tripling the channel count as the spatial resolution decreases, until a sufficient number of channels is reached (usually $128$ for local feature backbones~\cite{cit:disk, cit:aslfeat, cit:r2d2}). This strategy, marked by a triple rate increase in convolutional depth as spatial resolution halves, effectively redistributes the network's convolutional depth. It ensures minimal depth in early layers while compensating for the reduced parameter count accross the backbone. This approach not only significantly reduces the computational load in the early stages, particularly for high-resolution images, but also optimizes the network's overall capacity through more effective management of convolutional depth.
We found a good trade-off between spatial accuracy and speedup gains by starting with $C=4$  channels and concluding at $C=128$ in the final encoder block, achieving a spatial resolution of $\nicefrac{H}{32} \times \nicefrac{W}{32}$.




 
 Our network's simplicity is anchored in blocks called basic layers, a 2D convolution with kernel sizes from $1$ to $3$, ReLU + BatchNorm, and a stride of 2 for resolution reduction, forming convolutional blocks, each a composite of basic layers. The backbone features six blocks, halving resolution and increasing depth in sequence: $\{4,8,24,64,64,128\}$, plus a fusion block for multi-resolution features. More details on architecture are in the supplementary material.

\subsection{Local Feature Extraction}
In this section, we describe how our backbone is used to extract local features and perform dense matches.

\vspace{-0.3cm}
\paragraph{Descriptor head.}
The descriptor head extracts a dense feature map $\mathbf{F} \in \mathbb{R}^{\nicefrac{H}{8} \times \nicefrac{W}{8} \times 64}$, obtained by merging multi-scale features from the encoder. By using a feature pyramid strategy~\cite{cit:fpn}, we inexpensively increase the receptive field of the network by applying successive convolution blocks until $1/32$ of original resolution is achieved, a strategy that has demonstrated success in local feature extraction  to increase robustness to viewpoint changes~\cite{cit:aslfeat, cit:alike, cit:disk} and a key ingredient for small network backbones to work well in practice. We merge the intermediate representation at three different scale levels:$\{\nicefrac{1}{8}, \nicefrac{1}{16}, \nicefrac{1}{32}\}$ by bilinearly upsampling and projecting all intermediate representations to $\nicefrac{H}{8} \times \nicefrac{W}{8} \times 64$ followed by element-wise summation. Finally, a convolutional fusion block composed of three basic layers is used to combine the representations into the final feature representation $\mathbf{F}$. 
 An additional convolutional block is used to regress a reliability map $\mathbf{R} \in \mathbb{R}^{\nicefrac{H}{8} \times \nicefrac{W}{8}}$, which models the unconditional probability $\mathbf{R}_{i,j}$ that a given local feature $\mathbf{F}_{i,j}$ can be matched confidently. An overview of our method is shown in~\cref{fig:method}.

\vspace{-0.3cm}
\paragraph{Keypoint head.}
In general, backbones for local feature extracion rely on UNets~\cite{cit:disk}, VGG~\cite{cit:superpoint}, and ResNets~\cite{cit:aslfeat}. The strategy used in SuperPoint~\cite{cit:superpoint} offers the fastest approach to extract pixel-level keypoints. It uses features in the final encoder with $1/8$ of the original image resolution, and extracts pixel-level keypoints by classifying the coordinate of the keypoint in a flattened $8 \times 8$ grid from the feature embeddings. We adopt a strategy similar to SuperPoint, but with a major difference. We introduce a novel approach that employs a dedicated parallel branch for keypoint detection focused on low-level image structures. As shown in the ablation experiments (\cref{sec:ablations}), by jointly training a descriptor and a keypoint regressor within a single neural network backbone significantly degrades the performance of semi-dense matching for compact CNN architectures. 

Our key insight lies in the efficient utilization of the low-level features through a minimalist convolutional branch. To maintain spatial resolution without sacrificing speed, we represent the input image as a 2D grid comprised of $8 \times 8$ pixels on each grid cell, and we reshape each cell into $64$-dimensional features. This representation preserves spatial granularity within individual cells, while exploiting rapid $1 \times 1$ convolutions for regressing keypoint coordinates. After four convolutional layers, we obtain a keypoint embedding $\mathbf{K} \in \mathbb{R}^{\nicefrac{H}{8} \times \nicefrac{W}{8} \times (64+1)}$ encoding the logits of keypoint distribution inside a cell $\mathbf{k}_{i,j} \in \mathbf{K}$, and classify the keypoint as one of the $64$ possible positions inside $\mathbf{k}_{i,j} \in \mathbb{R}^{65}$ plus a dustbin to consider the case where no keypoint is found~\cite{cit:superpoint}. During inference, the dustbin is discarded and the heatmap is re-interpreted as an $8 \times 8$ cell. \cref{fig:method} depicts the entire process of the Keypoint Head. 

\vspace{-0.3cm}
\paragraph{Dense matching.}

Recent research~\cite{cit:loftr, cit:aspanformer} has demonstrated the benefits of dense image region matching, improving coverage and robustness. Our work proposes a lightweight module for dense feature matching, differing from other detector-free methods in two ways. Firstly, we can control memory and compute footprint by selecting top-$K$ image regions according to their reliability score $\mathbf{R}_{i,j}$ and caching them for future matching. Secondly, we propose a simple and lightweight Multi-Layer Perceptron (MLP) to perform coarse-to-fine matching without high-resolution feature maps~\cite{cit:loftr,cit:silk}, enabling us to perform semi-dense matching in resource-constrained settings.


Given the dense local feature map $\mathbf{F}$, which is at $1/8$ of input spatial resolution, or a subset $\mathbf{F}_s \in \mathbf{F}$, we propose a simple refinement strategy to recover pixel-level offsets.
Let $\mathbf{f}_a \in \mathbf{F_1}$ and $\mathbf{f}_b \in \mathbf{F_2}$ be two matching features obtained by traditional nearest neighbor matching from an image pair $(\mathbf{I_1}, \mathbf{I_2})$. We predict offsets $\mathbf{o} = \text{MLP}(\text{concat}(\mathbf{f}_a, \mathbf{f}_b)) \label{eqn:finematch}$, classifying the offset $(x,y)$ that leads to the correct pixel-level match at original image resolution: 
\begin{equation}
    (x, y) = \operatorname*{arg\,max}_{\substack{i \in \{1, \ldots, 8\} \\ j \in \{1, \ldots, 8\}}} \mathbf{o}(i,j), 
\end{equation}
\noindent where $\mathbf{o} \in \mathbb{R}^{8 \times 8}$ has the logits of a probability distribution over the possible offsets.

\begin{figure}[tb]
	\centering
	\includegraphics[width=0.8\columnwidth]{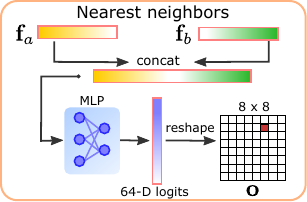}
	\caption{{\bf{Match refinement module for dense matching setting.}} This module learns to predict pixel-level offsets by only considering as input pairs of nearest neighbors from the original coarse-level features at $1/8$ of original spatial resolution, significantly saving memory and compute. }
	\label{fig:fine_match}
\end{figure}

The match refinement module is trained in an end-to-end manner alongside the backbone network, ensuring that the intermediate feature representation retains fine-grained spatial details within a compact embedding space. The offset prediction is conditioned on the coarsely matched feature pair $(\mathbf{f}_a, \mathbf{f}_b)$, reducing the search space. \cref{fig:fine_match} illustrates the lightweight match refinement module. 

\subsection{Network Training}

We train XFeat in a supervised manner with pixel-level ground truth correspondences. We assume image pairs $(\mathbf{I}_1,\mathbf{I}_2)$ with $N$ matching pixels $\mathbf{M}_{I_1 \leftrightarrow I_2 } \in \mathbb{R}^{N \times 4}$, where the first two columns of $\mathbf{M}_{I_1 \leftrightarrow I_2 }$ encode the $(x,y)$ coordinates of the points in $\mathbf{I}_1$, and the last two columns for $\mathbf{I}_2$.

\vspace{-0.3cm}
\paragraph{Learning local descriptors.}
To supervise the local feature embeddings $\mathbf{F}$, we employ the negative log-likelihood (NLL) loss.
Descriptor sets $\mathbf{F}_1$ and $\mathbf{F}_2$ are sampled from the dense maps $\mathbf{F}_{(\cdot,\cdot)}$, and each is represented in $\mathbb{R}^{N \times 64}$, comprising $N$ $64$-dimensional descriptors.
The $i$-th rows $\mathbf{F}_1(i,\cdot)$ and $\mathbf{F}_2(i,\cdot)$ correspond to two descriptors of the same point from $\mathbf{I}_1$ and $\mathbf{I}_2$ respectively.
Then, the similarity matrix $\mathbf{S} \in \mathbb{R}^{N \times N}$ is obtained by: $\mathbf{S} = \mathbf{F}_1 \mathbf{F}^{\mathsf{T}}_2.$
Given the symmetry of matching, we take both matching directions~\cite{cit:loftr}, resulting in the dual-softmax loss $\mathcal{L}_{ds}$, where the similarity measure of corresponding features lie in the main diagonal $\mathbf{S}_{ii}$ of $\mathbf{S}$ and $\text{softmax}_r$ is performed row-wise:
\begin{align}
\mathcal{L}_{ds} &= - \sum_{i} \log(\text{softmax}_r(\mathbf{S})_{ii}) \label{eqn:ds}  \nonumber\\
        &\quad - \sum_{i} \log(\text{softmax}_r(\mathbf{S}^{\mathsf{T}})_{ii}).
\end{align}

\vspace{-0.3cm}
\paragraph{Learning reliability.}
We supervise the reliability map during training by interpreting the dual-softmax probability as a confidence measure, denoted as $\bar{\mathbf{R}} \in \mathbb{R}^{N}$. $\bar{\mathbf{R}}_1$ and $\bar{\mathbf{R}}_2$ are obtained by matching $\mathbf{F}_1$ and $\mathbf{F}_2$ with the dual-softmax strategy: $\bar{\mathbf{R}}_1 = \text{max}_r(\text{softmax}_r(\mathbf{S}))$, and $\bar{\mathbf{R}}_2 = \text{max}_r(\text{softmax}_r(\mathbf{S}^{\mathsf{T}}))$, similarly to~\cref{eqn:ds}. As the training progresses, intuitively, distinct features will have high confidence matching probability. Thus, we supervise the reliability map directly with the L1 loss given the dual softmax scores $\bar{\mathbf{R}}_1$ and $\bar{\mathbf{R}}_2$: 
%
\begin{equation}
    \mathcal{L}_{rel} = |\sigma(\mathbf{R}_1) - \bar{\mathbf{R}}_1 \odot \bar{\mathbf{R}}_2| + 
     |\sigma(\mathbf{R}_2) - \bar{\mathbf{R}}_1 \odot \bar{\mathbf{R}}_2|,
\end{equation}
\noindent where $\sigma$ is the sigmoid activation function and $\odot$ the Hadamard product. Note that for the reliability loss $\mathcal{L}_{rel}$, we only backpropagate the gradients through $\mathbf{R}$. 

\vspace{-0.3cm}
\paragraph{Learning pixel offsets.}
The match refinement module is supervised with pixel-level offsets obtained from the ground-truth correspondences $\mathbf{M}_{I_1 \leftrightarrow I_2 }$ at the original input image resolution. We also employ the NLL loss over the logits $\mathbf{o}$ described in \cref{eqn:finematch}. During training, corresponding descriptors $\mathbf{F}_1(i,\cdot)$ and $\mathbf{F}_2(i,\cdot)$, together with their ground-truth offset $(\bar{x}, \bar{y})$ are obtained using $\mathbf{M}_{I_1 \leftrightarrow I_2 }(i,\cdot)$, and the fine matching loss $\mathcal{L}_{fine }$ becomes:
\begin{equation}
    \mathcal{L}_{fine} = - \sum_{i} \log(\text{softmax}\left(\mathbf{o}_i\right))_{\bar{y}_i, \bar{x}_i}.    
\end{equation}
\vspace{-0.3cm}
\paragraph{Learning keypoints.}
Our keypoint detection branch is minimalist by design. Whilst it is possible to supervise the keypoint head with existing keypoint losses~\cite{cit:alike, cit:r2d2, cit:disk}, we chose to employ knowledge distillation from a larger teacher network to facilitate its learning. We opted for ALIKE~\cite{cit:alike} keypoints obtained from its tiny backbone to supervise our model. This choice is strategic, as the smaller backbone tends to concentrate on lower-level image features like corners, lines, and blobs, aligning well with our designed detector branch, given its limited receptive field size of $8 \times 8$ pixels. 
Given the keypoint raw logit map $\mathbf{K} \in \mathbb{R}^{\nicefrac{H}{8} \times \nicefrac{W}{8} \times (64+1)}$, we map keypoint coordinates from the teacher network $(t_x, t_y)$ inside each cell $\mathbf{k}_{i,j} \in \mathbb{R}^{65}$ to linear index $t_{idx} = (t_x  + t_y * 8), \quad t_{idx} \in \{0, 1, ..., 63\}$. To supervise the dustbin, when no keypoint is detected inside a cell $\mathbf{k}_{i,j}$, we set $t_{idx} = 64$. During training, we set an upper limit of samples for the no keypoint case to avoid class imbalance.
Finally, the NLL loss is employed to compute the keypoint loss $\mathcal{L}_{kp}$:
\begin{equation}
    \mathcal{L}_{kp} = - \sum_{k} \log(\text{softmax}\left(\mathbf{k}_{i,j}\right))_{t_{idx}}.
\end{equation}
The final loss $\mathcal{L}$ is then a linear combination of all losses:
\begin{equation}
    \mathcal{L} = \alpha \mathcal{L}_{ds} + \beta \mathcal{L}_{rel} + \gamma \mathcal{L}_{fine} + \delta \mathcal{L}_{kp},
\end{equation}
\noindent where $\{\alpha, \beta, \gamma, \delta\}$ are hyperparameters to adjust the magnitude of the different losses.
\section{Experiments}
We evaluate XFeat on relative camera pose estimation, visual localization, and homography estimation. We also present ablations to justify our design decisions, and a comprehensive runtime analysis in a GPU-free setting.

\begin{table*}[tb!]
\centering
\caption{{\bf{Megadepth-1500 relative camera pose estimation.}} Our method achieves superior performance compared to other lightweight methods, while also outperforming SuperPoint at 9$\times$ speedup, and with comparable results to DISK at 16$\times$ speedup. $^*$ denotes 10k keypoints. FPS is the average of $30$ frames $\pm$ standard deviation computed in VGA resolution. Best in bold, second best underlined, separated by method class (standard/fast). \textsuperscript{+} indicates code used as provided by authors without hardware optimization.}
\label{tab:relative_pose}
\resizebox{0.91\textwidth}{!}{%
\begin{tabular}{@{}clccc|ccrrr@{}}
\toprule
& \textbf{Method} & \textbf{AUC@$5^\circ$} & \textbf{AUC@$10^\circ$} & \textbf{AUC@$20^\circ$} & \textbf{Acc@$10^\circ$} & \textbf{MIR} & \textbf{\#inliers} & \textbf{dim} & \textbf{FPS} \\
\midrule
\multirow{6}{*}{\rotatebox[origin=c]{90}{Standard}} 
& SiLK~\cite{cit:silk}         & 14.7 & 21.5 & 29.3 & 31.9  & 0.17 & 235 & \textbf{32-f} & {2.8 $\pm$ 0.08}  \\
& SiLK$^{*}$~\cite{cit:silk}         & 16.2 & 23.2 & 31.8 & 34.7 & 0.14 & 478 & \textbf{32-f} & \underline{2.9 $\pm$ 0.12} \\
& SuperPoint~\cite{cit:superpoint}   & 37.3 & 50.1 & 61.5 & \underline{67.4} & 0.35 & 495 & 256-f & \textbf{3.0 $\pm$ 0.07}  \\
& DISK~\cite{cit:disk}         & \underline{53.8} & \underline{65.9} & \underline{75.0} & \textbf{81.3} & \textbf{0.72} & \underline{1231} & \underline{128-f} & $1.2 \pm 0.01$  \\
& DISK$^{*}$~\cite{cit:disk}   & \textbf{55.2} & \textbf{66.8} & \textbf{75.3} & \textbf{81.3} & \underline{0.71} & \textbf{1997} & \underline{128-f} & $1.2 \pm 0.01$  \\
\hdashline \noalign{\vskip 0.5ex}
\multirow{4}{*}{\rotatebox[origin=c]{90}{Fast}} 
& ORB~\cite{cit:orb}  & 17.9 & 27.6 & 39.0 & 43.1 & 0.25 & 288 & \textbf{256-b} & \textbf{44.3 $\pm$ 1.18}  \\
& ZippyPoint~\cite{cit:zippypoint}   & 23.6 & 34.9 & 46.3 & 51.8 & 0.23 & 192 & \textbf{256-b} & \textsuperscript{+}$1.8 \pm 0.06$  \\
& ALIKE~\cite{cit:alike}        & \underline{49.4} & \underline{61.8} & \underline{71.4} & \underline{77.7} & 0.47 & 333 & \underline{64-f} & {5.3 $\pm$ 0.33}  \\
& XFeat         & {42.6} & {56.4} & {67.7} & {74.9} & \underline{0.55} & \underline{892} & \underline{64-f} & \underline{$27.1 \pm 0.33$}  \\
& XFeat$^{*}$   & \textbf{50.2} & \textbf{65.4} & \textbf{77.1} & \textbf{85.1} & \textbf{0.74} & \textbf{1885} & \underline{64-f} & $19.2 \pm 1.12$  \\
\bottomrule
\end{tabular}
}
\end{table*}

\noindent\textbf{Training.} XFeat was implemented on PyTorch~\cite{cit:pytorch} and trained on a blend of Megadepth~\cite{cit:megadepth} and synthetically warped COCO~\cite{cit:coco} images, using a 6:4 ratio, with images resized to $(W=800, H=600)$. Hybrid training was found to enhance generalization in our experiments (\cref{sec:ablations}), aligning with recent findings~\cite{cit:lightglue}. The training involved batches of $10$ image pairs using the Adam optimizer~\cite{cit:adam}, leading to convergence within $36$ hours on an NVIDIA RTX 4090 GPU. Further details on computational resource utilization and hyperparameter specifics are provided in the supplementary material.


 \noindent\textbf{XFeat inference.} We considered two settings: Sparse (XFeat) and semi-dense matching (XFeat$^{*}$), both utilizing the same pretrained backbone. In XFeat, we extracted up to $4{,}096$ keypoints from the keypoint heatmap $\mathbf{K}$, using their scores derived from the keypoint and reliability confidences: $score = \mathbf{K}_{i,j} \cdot \mathbf{R}_{i,j}$. Local features were then bicubically interpolated from $\mathbf{F}$ at these keypoint locations and matched with Mutual Nearest Neighbor (MNN) search. For XFeat$^{*}$, we enhanced features by processing images at 2 different scales ($0.65$ and $1.3$, resizing the image internally after receiving the input), retaining the top $10{,}000$ features according to their reliability. We used MNN search and offset refinement to match the features, retaining only those with offset prediction confidence above $0.2$. 


\noindent\textbf{Baselines.} Among the selected baselines, DISK~\cite{cit:disk} sets a high benchmark in accuracy at the cost of increased computational demand. This is followed by SiLK~\cite{cit:silk}, SuperPoint~\cite{cit:superpoint}, ZippyPoint~\cite{cit:superpoint}, and ALIKE~\cite{cit:alike}. For SiLK and ALIKE, we opted for their smallest available backbones -- \textit{ALIKE-Tiny} and \textit{VGGnp-$\mu$} -- aligning with our focus on models emphasizing compute efficiency. Finally, ORB~\cite{cit:alike} represents the upper limit in terms of speed.
Thus, we evaluate XFeat against the current state-of-the-art through a diverse set of baselines covering the spectrum of computational expense and accuracy.
We use the top $4,096$ detected keypoints for all baselines, except for those marked with $^{*}$, where the top $10,000$ keypoints are used. For matching, MNN search is employed. ZippyPoint model was used in its form as provided by the authors without hardware-specific compilation, due to the lack of clear instructions.

\subsection{Relative pose estimation}

\begin{table}[tb!]
\centering
\caption{\new{{\bf{ScanNet-1500 relative pose estimation.}} XFeat and XFeat$^{*}$ exhibit better generalization to indoor scenes.}}
\label{tab:indoor_eval}
\resizebox{0.99\columnwidth}{!}{%
\begin{tabular}{@{}rccc|ccc@{}}
\toprule
\multirow{2}{*}{\textbf{AUC}} & \textbf{Super-} & \textbf{DISK} & \multirow{2}{*}{\textbf{ORB}} & \multirow{2}{*}{\textbf{ALIKE}} & \textbf{XFeat/} \\
  & \textbf{Point} & ($4$k/$10$k) & & & \textbf{XFeat}$^{*}$  \\ 
\midrule 
@$5^\circ$      & {12.5} & {\phantom{0}9.6} / {11.3} & {\phantom{0}9.0} & {\phantom{0}8.0} & \underline{16.7} / {\textbf{18.4}}  \\
@$10^\circ$     & {24.4} & {19.3} / {22.3} & {18.5} & {16.4} & \underline{32.6} / {\textbf{34.7}}  \\
@$20^\circ$     & {36.7} & {30.4} / {33.9} & {29.9} & {25.9} & \underline{47.8} / {\textbf{50.3}}  \\
\bottomrule
\end{tabular}
}
\vspace{-0.2cm}
\end{table}

\begin{figure*}[tb]
	\centering
	\includegraphics[width=0.91\textwidth]{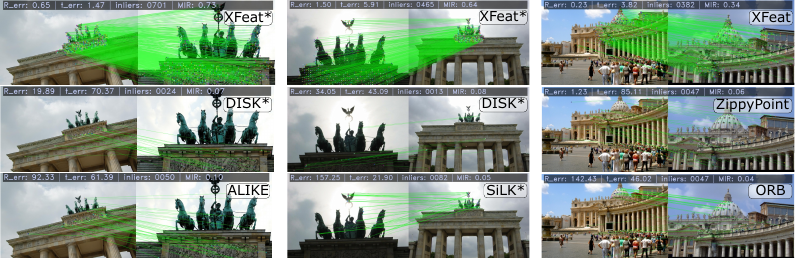}
	\caption{{\bf{Qualitative results on Megadepth-1500.}} XFeat$^{*}$ and XFeat demonstrate exceptional robustness against variations in viewpoint and illumination. This is especially evident in challenging scenarios where heavy methods like DISK$^{*}$ breaks and XFeat$^{*}$ provide accurate relative pose $16 \times$ times faster in semi-dense settings with a comparable number of local features. }
	\label{fig:qualitative}
\end{figure*}

\noindent\textbf{Setup.} Megadepth~\cite{cit:megadepth} \new{and ScanNet~\cite{cit:scannet}} test sets are used as in previous works~\cite{cit:loftr, cit:lightglue}, providing camera poses on scenes that do not overlap with our training set. The scenes contain significant viewpoint and illumination changes simultaneously and present repetitive structures, posing a significant challenge. LO-RANSAC~\cite{cit:poselib} is used to estimate the essential matrix. We search for the optimal threshold for each method, and resize the images such that the maximum dimension becomes $1{,}200$ pixels for Megadepth \new{and use the default (VGA) resolution for ScanNet}.

\noindent\textbf{Metrics.} We use the area under the curve (AUC) at thresholds of $\{5^{\circ},10^\circ,20^\circ\}$~\cite{cit:loftr, cit:lightglue}. Additionally, we report the Acc@$10^\circ$, which is the proportion of poses where the maximum angular error is below $10$ degrees, the mean inlier ratio (MIR), which is the ratio of matching points that comply with the estimated model after RANSAC, and the number of inlier points (\#inliers). Finally, we measure the frames per second (FPS) of each method on a budget-friendly laptop without GPU and an Intel(R) i5-1135G7 @ 2.40GHz CPU. We also indicate whether the descriptor is floating-point (denoted by \textbf{f}) or binary-based (denoted by \textbf{b}) and report the descriptor dimensionality (dim).


\noindent\textbf{Results.} \label{sec:results_pose} \cref{tab:relative_pose} shows the metrics on the relative camera pose estimation task \new{on Megadepth-1500}.
Our method is much faster (5$\times$) than the fastest available learning-based solution (ALIKE) and achieves competitive results in the sparse setting on several metrics. Moreover, it can deliver state-of-the-art results for the dense matching configuration using $10,000$ descriptors on AUC@$20^\circ$, Acc@$10^\circ$ and MIR in a fair comparison with DISK${^*}$, a much heavier model, considering the same number of descriptors. \cref{fig:qualitative} shows examples where XFeat stands out over existing solutions. Our method also allows more efficient matching with low-dimensional descriptors (64-f) compared to DISK and SuperPoint. \new{Detailed timing analysis is provided in the supplementary material alongside additional quantitative comparison with recent popular learned matchers~\cite{cit:lightglue, cit:patch2pix, cit:loftr}}. It is worth mentioning that we obtain state-of-the-art results in more loose thresholds due to the requirement of interpolating the descriptors and predicting offsets at coarser resolution. \new{\cref{tab:indoor_eval} shows AUC values for the most competitive methods in ScanNet-1500 indoor imagery. Notice that none of the methods were retrained. DISK and ALIKE show signs of bias towards landmark datasets, while our approach demonstrates superior generalization. A more detailed discussion and qualitative results for ScanNet and Megadepth are available in the supplementary material.} 

\subsection{Homography estimation}
\noindent\textbf{Setup.} We used the widely adopted HPatches~\cite{cit:hpatches} dataset containing sequences of images from planar scenes with moderate to strong viewpoint and illumination changes. Similarly to relative pose estimation, we used MAGSAC++~\cite{cit:magsac} to robustly estimate the homography transformation given the correspondences of each method.

\paragraph{Metrics.} We followed ALIKE~\cite{cit:alike} protocol and estimated Mean Homography Accuracy (MHA).  We used predefined thresholds of $\{3, 5, 7\}$ pixels. The accuracy was computed considering the average corner error in pixels by warping the four reference image corners to target images using the ground-truth homography and estimated one.


\noindent\textbf{Results.} \cref{table:homography} shows that our method is on par with the most accurate descriptors, reinforcing the robustness of our proposed keypoint and descriptor heads. In contrast, the performance of other lightweight solutions as ORB and SiLK are heavily compromised on the illumination and viewpoint splits, due to their limited capacity in handling agressive viewpoint and illumination changes present in the hardest image pairs. Our method also stands out for less strict thresholds, as discussed in \cref{sec:results_pose} -- Results.
\begin{table}[tb!]
	\centering
	\caption{{\bf Homography estimation on HPatches.} All methods perform well due to RANSAC except ORB and SiLK which break on several illumination sequences. XFeat provides high quality homography estimation with a fraction of compute. Best in bold, second best underlined, separated by standard and fast methods.}
    \resizebox{0.99\columnwidth}{!}{%
	\begin{tabular}{lcccccc}
		\toprule      
		\multirow{3}{*}{\textbf{Method}} &\multicolumn{3}{c}{\textbf{Illumination}} & \multicolumn{3}{c}{\textbf{Viewpoint}}   \\ 
                                         & \multicolumn{3}{c}{MHA}    &   \multicolumn{3}{c}{MHA}    \\
                                         & @3 & @5 & @7     &      @3 & @5 & @7  \\
		\cmidrule(l){1-1}   \cmidrule(lr){2-4} \cmidrule(l){5-7}
        SiLK     & \underline{78.5} & 82.3 & 83.8 & 48.6 & 59.6 & 62.5 \\
        SuperPoint     & \textbf{94.6} & \underline{98.5} & \underline{98.8} & \textbf{71.1} & \textbf{79.6} & \textbf{83.9} \\
        DISK     & \textbf{94.6} & \textbf{98.8} & \textbf{99.6} & \underline{66.4} & \underline{77.5} & \underline{81.8} \\
        \hdashline \noalign{\vskip 0.5ex}
        ORB     & 74.6 & 84.6 & 85.4 & 63.2 & 71.4 & 78.6 \\
        ZippyPoint     & 94.2 & 96.9 & 98.5 & 66.1 & 76.8 & 80.7 \\
        ALIKE & \underline{94.6} & \textbf{98.5} & \textbf{99.6} & \underline{68.2} & \underline{77.5} & \underline{81.4} \\
        XFeat     & \textbf{95.0} & \underline{98.1} & \underline{98.8} & \textbf{68.6} & \textbf{81.1} & \textbf{86.1} \\
		\bottomrule		
	\end{tabular}  
    }
	\label{table:homography}
\end{table}

\subsection{Visual localization}

\noindent\textbf{Setup.} The hierarchical localization pipeline HLoc~\cite{cit:hloc} is used to localize images of day and night scenes from the Aachen dataset~\cite{cit:aachen}. Given the provided keypoint correspondences, HLoc triangulates an SfM map using the available ground-truth camera poses. A separate set of query images are then localized within the 3D map using the keypoint matches. For a fair comparison, we resize the images such that maximum dimension is held at $1,024$ pixels, and extract the top $4,096$ keypoints for all approaches. 

\paragraph{Metrics.} We use the standard metric provided by HLoc, which is the accuracy of correctly estimated camera poses within thresholds of position errors $\{0.25m, 0.5m, 5m\}$ and rotation errors $\{2^{\circ}, 5^{\circ}, 10^{\circ}\}$ respectively. 

 
\begin{table}[tb!]
	\centering \vspace{0.2cm}
	\caption{{\bf Visual localization on Aachen day-night.} XFeat enables fast and accurate localization, especially on the more challenging case of matching day-to-night images, being on-par with the state-of-the-art on thresholds above $0.5m, 5^{\circ}$. Best in bold, second best underlined, separated by standard and fast methods.}
    \resizebox{0.99\columnwidth}{!}{%
	\begin{tabular}{lcccccc}
		\toprule      
		\multirow{3}{*}{\textbf{Method}} &\multicolumn{3}{c}{\textbf{Day}} & \multicolumn{3}{c}{\textbf{Night}}   \\ 
                                         & $0.25m$ & $0.5m$ & $5m$     &     $0.25m$ & $0.5m$ & $5m$  \\
                                         & $2^{\circ}$ & $5^{\circ}$ & $10^{\circ}$     &   $2^{\circ}$ & $5^{\circ}$ & $10^{\circ}$  \\
		\cmidrule(l){1-1}   \cmidrule(lr){2-4} \cmidrule(l){5-7}
        SuperPoint      & \textbf{87.4} & \underline{93.2} & \underline{97.0} & \underline{77.6} & \underline{85.7} & \underline{95.9} \\
        DISK            & \underline{86.9} & \textbf{95.1} & \textbf{97.8} & \textbf{83.7} & \textbf{89.8} & \textbf{99.0} \\
        \hdashline \noalign{\vskip 0.5ex}
        ORB             & 66.9 & 76.1 & 81.7 & 10.2 & 12.2 & 19.4 \\
        ZippyPoint      & 80.7 & 88.6 & 93.7 & 61.2 & 70.4 & 79.6 \\
    	ALIKE           & \textbf{85.7} & \textbf{92.4} & \textbf{96.7} & \textbf{81.6} & 88.8 & \textbf{99.0} \\
        XFeat           & \underline{84.7}  & \underline{91.5} & \underline{96.5} & \underline{77.6} & \textbf{89.8} & \underline{98.0} \\
		\bottomrule		
	\end{tabular}  
    }
	\label{table:vloc}
\end{table}

\paragraph{Results.} Table~\ref{table:vloc} presents the results of the visual localization experiment. Our method demonstrates similar performance to leading approaches as SuperPoint and DISK, while achieving a significant speed advantage, being at least $9$ times faster and with a more compact descriptor. These findings challenge the prevailing trend in the literature to employ large and more intricate models for downstream tasks. Contrarily, they underscore the efficacy of simpler models that not only match accuracy but also offer the benefits of efficient operation on resource-constrained systems.

\subsection{Ablation} \label{sec:ablations}

    \begin{table}[tb!]
    \centering
    \caption{{\bf{Ablation on Megadepth-1500.}} We ablate the architecture and training strategies for relative camera pose estimation.} \label{tab:ablation}
    \begin{tabular}{lcc} 
    \toprule
    \textbf{Strategy } & \multicolumn{2}{c} {\textbf{AUC@$5^\circ$}}  \\
    & XFeat & XFeat$^{*}$ \\
    \midrule
    Default                          & 42.6  & 50.2 \\
    (i) No synthetic data            & 41.5  & 33.9 \\
    (ii) Smaller model               & 37.4  & 40.7 \\
    (iii) Joint keypoint extraction  & 42.9  & 39.7 \\
    (iv) No match refinement         & -     & 38.6 \\
    \bottomrule
    \end{tabular}
    \vspace{-0.4cm}
    \end{table}


We ablate our architecture in several configurations, which are listed in \cref{tab:ablation}. We evaluate whether training with additional synthetic warps (i) may help the model to become more robust to challenging image pairs. Training on both real and synthetic warps is beneficial, especially for the dense matching setting. Second, we evaluate if we can further reduce channel count in the network (ii). We halve the channels of the last three convolutional blocks to $32$ instead of $64$, but performance significantly degrades for both sparse and dense settings. 
We also demonstrate the rationale behind devising a parallel branch for keypoint detection. Without the proposed keypoint head (iii), an additional convolutional block is used on top of the output descriptor embeddings akin to SuperPoint. As shown in~\cref{tab:ablation}, XFeat$^{*}$ experiences degradation in performance when trained under this specific setup, since the limited network size constrains the capacity of intermediate embeddings, rendering them less effective for semi-dense matching in non-repeatable regions, adversely affecting the match refinement task. Thus, we opted to design a parallel branch which offers great trade-off between sparse and dense matching as shown in ~\cref{tab:ablation} -- Default and (iii). Lastly, we evaluate the benefits of our proposed match refinement module. For XFeat$^{*}$, the match refinement step is critical for enhancing accuracy. In our benchmarks, this module incurs only an additional $11\%$ inference cost compared to MNN matching for an average of $10{,}000$ descriptors. \new{Please check the supplementary material for a thorough timing analysis.}


\section{Conclusion}
This work introduced XFeat, a lightweight CNN architecture for accelerated feature extraction, applicable to both sparse and semi-dense image matching. With experiments on three different tasks and ablation analyses, we showed that it is possible to achieve fast and accurate image matching without resorting to advanced low-level hardware optimizations. This stands in contrast to the prevalent trend of deploying increasingly large and convoluted models. We believe XFeat paves the way for next-generation applications in augmented reality and mobile robotics, where efficient and general data-driven solutions remain crucial for real-world deployment, particularly in mobile applications.

\new{
\vspace{-0.15in}
\paragraph{Acknowledgements.} This work was supported by CAPES, CNPq, FAPEMIG, Google and ANR (ANR-23-CE23-0003-01), to whom we are grateful. 
}

{
    \small
    \bibliographystyle{ieeenat_fullname}
    \bibliography{main}

\begin{thebibliography}{47}
\providecommand{\natexlab}[1]{#1}
\providecommand{\url}[1]{\texttt{#1}}
\expandafter\ifx\csname urlstyle\endcsname\relax
  \providecommand{\doi}[1]{doi: #1}\else
  \providecommand{\doi}{doi: \begingroup \urlstyle{rm}\Url}\fi

\bibitem[Aiger et~al.(2023)Aiger, Araujo, and Lynen]{cit:cann}
D. Aiger, A. Araujo, and S. Lynen.
\newblock {Yes, we CANN: Constrained Approximate Nearest Neighbors for local
  feature-based visual localization}.
\newblock In \emph{ICCV}, 2023.

\bibitem[Balntas et~al.(2017)Balntas, Lenc, Vedaldi, and
  Mikolajczyk]{cit:hpatches}
Vassileios Balntas, Karel Lenc, Andrea Vedaldi, and Krystian Mikolajczyk.
\newblock Hpatches: A benchmark and evaluation of handcrafted and learned local
  descriptors.
\newblock In \emph{CVPR}, pages 5173--5182, 2017.

\bibitem[Barath et~al.(2020)Barath, Noskova, Ivashechkin, and
  Matas]{cit:magsac}
Daniel Barath, Jana Noskova, Maksym Ivashechkin, and Jiri Matas.
\newblock Magsac++, a fast, reliable and accurate robust estimator.
\newblock In \emph{CVPR}, pages 1304--1312, 2020.

\bibitem[Barroso-Laguna et~al.(2019)Barroso-Laguna, Riba, Ponsa, and
  Mikolajczyk]{cit:keynet}
Axel Barroso-Laguna, Edgar Riba, Daniel Ponsa, and Krystian Mikolajczyk.
\newblock Key. net: Keypoint detection by handcrafted and learned cnn filters.
\newblock In \emph{ICCV}, pages 5836--5844, 2019.

\bibitem[Chen et~al.(2022)Chen, Luo, Zhou, Tian, Zhen, Fang, Mckinnon, Tsin,
  and Quan]{cit:aspanformer}
Hongkai Chen, Zixin Luo, Lei Zhou, Yurun Tian, Mingmin Zhen, Tian Fang, David
  Mckinnon, Yanghai Tsin, and Long Quan.
\newblock Aspanformer: Detector-free image matching with adaptive span
  transformer.
\newblock In \emph{ECCV}, pages 20--36. Springer, 2022.

\bibitem[Dai et~al.(2017)Dai, Chang, Savva, Halber, Funkhouser, and
  Nie{\ss}ner]{cit:scannet}
Angela Dai, Angel~X Chang, Manolis Savva, Maciej Halber, Thomas Funkhouser, and
  Matthias Nie{\ss}ner.
\newblock Scannet: Richly-annotated 3d reconstructions of indoor scenes.
\newblock In \emph{CVPR}, pages 5828--5839, 2017.

\bibitem[DeTone et~al.(2018)DeTone, Malisiewicz, and
  Rabinovich]{cit:superpoint}
Daniel DeTone, Tomasz Malisiewicz, and Andrew Rabinovich.
\newblock Superpoint: Self-supervised interest point detection and description.
\newblock In \emph{CVPRW}, pages 224--236, 2018.

\bibitem[Edstedt et~al.(2023)Edstedt, Athanasiadis, Wadenb{\"a}ck, and
  Felsberg]{cit:dkm}
Johan Edstedt, Ioannis Athanasiadis, M{\aa}rten Wadenb{\"a}ck, and Michael
  Felsberg.
\newblock Dkm: Dense kernelized feature matching for geometry estimation.
\newblock In \emph{CVPR}, pages 17765--17775, 2023.

\bibitem[Gleize et~al.(2023)Gleize, Wang, and Feiszli]{cit:silk}
Pierre Gleize, Weiyao Wang, and Matt Feiszli.
\newblock Silk: Simple learned keypoints.
\newblock In \emph{ICCV}, pages 22499--22508, 2023.

\bibitem[Harris et~al.(1988)Harris, Stephens, et~al.]{cit:harris}
Chris Harris, Mike Stephens, et~al.
\newblock A combined corner and edge detector.
\newblock In \emph{Alvey vision conference}, pages 10--5244. Citeseer, 1988.

\bibitem[He et~al.(2016)He, Zhang, Ren, and Sun]{cit:resnet}
Kaiming He, Xiangyu Zhang, Shaoqing Ren, and Jian Sun.
\newblock Deep residual learning for image recognition.
\newblock In \emph{CVPR}, pages 770--778, 2016.

\bibitem[Howard et~al.(2017)Howard, Zhu, Chen, Kalenichenko, Wang, Weyand,
  Andreetto, and Adam]{cit:mobilenet}
Andrew~G Howard, Menglong Zhu, Bo Chen, Dmitry Kalenichenko, Weijun Wang,
  Tobias Weyand, Marco Andreetto, and Hartwig Adam.
\newblock Mobilenets: Efficient convolutional neural networks for mobile vision
  applications.
\newblock \emph{arXiv:1704.04861}, 2017.

\bibitem[Kanakis et~al.(2023)Kanakis, Maurer, Spallanzani, Chhatkuli, and
  Van~Gool]{cit:zippypoint}
Menelaos Kanakis, Simon Maurer, Matteo Spallanzani, Ajad Chhatkuli, and Luc
  Van~Gool.
\newblock Zippypoint: Fast interest point detection, description, and matching
  through mixed precision discretization.
\newblock In \emph{CVPRW}, pages 6113--6122, 2023.

\bibitem[Karpur et~al.(2024)Karpur, Perrotta, Martin-Brualla, Zhou, and
  Araujo]{cit:lfm3d}
A. Karpur, G. Perrotta, R. Martin-Brualla, H. Zhou, and A. Araujo.
\newblock {LFM-3D: Learnable Feature Matching Across Wide Baselines Using 3D
  Signals}.
\newblock In \emph{3DV}, 2024.

\bibitem[Kingma and Ba(2015)]{cit:adam}
Diederik~P. Kingma and Jimmy Ba.
\newblock Adam: {A} method for stochastic optimization.
\newblock In \emph{ICLR}, 2015.

\bibitem[Larsson and contributors(2020)]{cit:poselib}
Viktor Larsson and contributors.
\newblock {PoseLib - Minimal Solvers for Camera Pose Estimation}, 2020.

\bibitem[Li and Snavely(2018)]{cit:megadepth}
Zhengqi Li and Noah Snavely.
\newblock Megadepth: Learning single-view depth prediction from internet
  photos.
\newblock In \emph{CVPR}, pages 2041--2050, 2018.

\bibitem[Lin et~al.(2014)Lin, Maire, Belongie, Hays, Perona, Ramanan,
  Doll{\'a}r, and Zitnick]{cit:coco}
Tsung-Yi Lin, Michael Maire, Serge Belongie, James Hays, Pietro Perona, Deva
  Ramanan, Piotr Doll{\'a}r, and C~Lawrence Zitnick.
\newblock Microsoft coco: Common objects in context.
\newblock In \emph{ECCV}, pages 740--755. Springer, 2014.

\bibitem[Lin et~al.(2017)Lin, Doll{\'a}r, Girshick, He, Hariharan, and
  Belongie]{cit:fpn}
Tsung-Yi Lin, Piotr Doll{\'a}r, Ross Girshick, Kaiming He, Bharath Hariharan,
  and Serge Belongie.
\newblock Feature pyramid networks for object detection.
\newblock In \emph{CVPR}, pages 2117--2125, 2017.

\bibitem[Lindenberger et~al.(2023)Lindenberger, Sarlin, and
  Pollefeys]{cit:lightglue}
Philipp Lindenberger, Paul-Edouard Sarlin, and Marc Pollefeys.
\newblock {LightGlue: Local Feature Matching at Light Speed}.
\newblock In \emph{ICCV}, 2023.

\bibitem[Lowe(2004)]{cit:sift}
David~G Lowe.
\newblock Distinctive image features from scale-invariant keypoints.
\newblock \emph{IJCV}, 60:\penalty0 91--110, 2004.

\bibitem[Luo et~al.(2020)Luo, Zhou, Bai, Chen, Zhang, Yao, Li, Fang, and
  Quan]{cit:aslfeat}
Zixin Luo, Lei Zhou, Xuyang Bai, Hongkai Chen, Jiahui Zhang, Yao Yao, Shiwei
  Li, Tian Fang, and Long Quan.
\newblock Aslfeat: Learning local features of accurate shape and localization.
\newblock In \emph{CVPR}, pages 6589--6598, 2020.

\bibitem[Mishchuk et~al.(2017)Mishchuk, Mishkin, Radenovic, and
  Matas]{cit:hardnet}
Anastasiia Mishchuk, Dmytro Mishkin, Filip Radenovic, and Jiri Matas.
\newblock Working hard to know your neighbor's margins: Local descriptor
  learning loss.
\newblock \emph{NeurIPS}, 30, 2017.

\bibitem[Mishkin et~al.(2018)Mishkin, Radenovic, and Matas]{cit:affnet}
Dmytro Mishkin, Filip Radenovic, and Jiri Matas.
\newblock Repeatability is not enough: Learning affine regions via
  discriminability.
\newblock In \emph{ECCV}, pages 284--300, 2018.

\bibitem[Mur-Artal and Tard{\'o}s(2017)]{cit:orbslam}
Raul Mur-Artal and Juan~D Tard{\'o}s.
\newblock Orb-slam2: An open-source slam system for monocular, stereo, and
  rgb-d cameras.
\newblock \emph{IEEE Trans. on Robotics.}, 33\penalty0 (5):\penalty0
  1255--1262, 2017.

\bibitem[Nascimento et~al.(2019)Nascimento, Potje, Martins, Cadar, Campos, and
  Bajcsy]{cit:geobit}
Erickson~R Nascimento, Guilherme Potje, Renato Martins, Felipe Cadar, Mario~FM
  Campos, and Ruzena Bajcsy.
\newblock Geobit: A geodesic-based binary descriptor invariant to non-rigid
  deformations for rgb-d images.
\newblock In \emph{ICCV}, pages 10004--10012, 2019.

\bibitem[Noh et~al.(2017)Noh, Araujo, Sim, Weyand, and Han]{cit:delf}
Hyeonwoo Noh, Andre Araujo, Jack Sim, Tobias Weyand, and Bohyung Han.
\newblock Large-scale image retrieval with attentive deep local features.
\newblock In \emph{ICCV}, pages 3456--3465, 2017.

\bibitem[Paszke et~al.(2019)Paszke, Gross, Massa, Lerer, Bradbury, Chanan,
  Killeen, Lin, Gimelshein, Antiga, et~al.]{cit:pytorch}
Adam Paszke, Sam Gross, Francisco Massa, Adam Lerer, James Bradbury, Gregory
  Chanan, Trevor Killeen, Zeming Lin, Natalia Gimelshein, Luca Antiga, et~al.
\newblock Pytorch: An imperative style, high-performance deep learning library.
\newblock \emph{NeurIPS}, 32, 2019.

\bibitem[Potje et~al.(2017)Potje, Resende, Campos, and
  Nascimento]{cit:potje_sfm}
Guilherme Potje, Gabriel Resende, Mario Campos, and Erickson~R Nascimento.
\newblock Towards an efficient 3d model estimation methodology for aerial and
  ground images.
\newblock \emph{Mach. Vis. and Applications.}, 28:\penalty0 937--952, 2017.

\bibitem[Potje et~al.(2021)Potje, Martins, Chamone, and Nascimento]{cit:deal}
Guilherme Potje, Renato Martins, Felipe Chamone, and Erickson Nascimento.
\newblock Extracting deformation-aware local features by learning to deform.
\newblock \emph{NeurIPS}, 34:\penalty0 10759--10771, 2021.

\bibitem[Potje et~al.(2022)Potje, Martins, Cadar, and Nascimento]{cit:geopatch}
Guilherme Potje, Renato Martins, Felipe Cadar, and Erickson~R Nascimento.
\newblock Learning geodesic-aware local features from rgb-d images.
\newblock \emph{CVIU}, 219:\penalty0 103409, 2022.

\bibitem[Potje et~al.(2023)Potje, Cadar, Araujo, Martins, and
  Nascimento]{cit:dalf}
Guilherme Potje, Felipe Cadar, Andr{\'e} Araujo, Renato Martins, and Erickson~R
  Nascimento.
\newblock Enhancing deformable local features by jointly learning to detect and
  describe keypoints.
\newblock In \emph{CVPR}, pages 1306--1315, 2023.

\bibitem[Revaud et~al.(2019)Revaud, Weinzaepfel, de~Souza, and
  Humenberger]{cit:r2d2}
Jerome Revaud, Philippe Weinzaepfel, C{\'{e}}sar~Roberto de Souza, and Martin
  Humenberger.
\newblock {R2D2:} repeatable and reliable detector and descriptor.
\newblock \emph{NeurIPS}, 2019.

\bibitem[Rublee et~al.(2011)Rublee, Rabaud, Konolige, and Bradski]{cit:orb}
Ethan Rublee, Vincent Rabaud, Kurt Konolige, and Gary Bradski.
\newblock Orb: An efficient alternative to sift or surf.
\newblock In \emph{ICCV}, pages 2564--2571. Ieee, 2011.

\bibitem[Sarlin et~al.(2019)Sarlin, Cadena, Siegwart, and Dymczyk]{cit:hloc}
Paul-Edouard Sarlin, Cesar Cadena, Roland Siegwart, and Marcin Dymczyk.
\newblock From coarse to fine: Robust hierarchical localization at large scale.
\newblock In \emph{CVPR}, pages 12716--12725, 2019.

\bibitem[Sarlin et~al.(2020)Sarlin, DeTone, Malisiewicz, and
  Rabinovich]{cit:superglue}
Paul-Edouard Sarlin, Daniel DeTone, Tomasz Malisiewicz, and Andrew Rabinovich.
\newblock Superglue: Learning feature matching with graph neural networks.
\newblock In \emph{CVPR}, pages 4938--4947, 2020.

\bibitem[Sattler et~al.(2018)Sattler, Maddern, Toft, Torii, Hammarstrand,
  Stenborg, Safari, Okutomi, Pollefeys, Sivic, et~al.]{cit:aachen}
Torsten Sattler, Will Maddern, Carl Toft, Akihiko Torii, Lars Hammarstrand,
  Erik Stenborg, Daniel Safari, Masatoshi Okutomi, Marc Pollefeys, Josef Sivic,
  et~al.
\newblock Benchmarking 6dof outdoor visual localization in changing conditions.
\newblock In \emph{CVPR}, pages 8601--8610, 2018.

\bibitem[Schonberger and Frahm(2016)]{cit:colmap}
Johannes~L Schonberger and Jan-Michael Frahm.
\newblock Structure-from-motion revisited.
\newblock In \emph{CVPR}, pages 4104--4113, 2016.

\bibitem[Simonyan and Zisserman(2014)]{cit:vgg}
Karen Simonyan and Andrew Zisserman.
\newblock Very deep convolutional networks for large-scale image recognition.
\newblock \emph{arXiv:1409.1556}, 2014.

\bibitem[Sun et~al.(2021)Sun, Shen, Wang, Bao, and Zhou]{cit:loftr}
Jiaming Sun, Zehong Shen, Yuang Wang, Hujun Bao, and Xiaowei Zhou.
\newblock Loftr: Detector-free local feature matching with transformers.
\newblock In \emph{CVPR}, pages 8922--8931, 2021.

\bibitem[Truong et~al.(2023)Truong, Danelljan, Timofte, and
  Van~Gool]{cit:pdcnet}
Prune Truong, Martin Danelljan, Radu Timofte, and Luc Van~Gool.
\newblock Pdc-net+: Enhanced probabilistic dense correspondence network.
\newblock \emph{IEEE TPAMI}, 2023.

\bibitem[Tyszkiewicz et~al.(2020)Tyszkiewicz, Fua, and Trulls]{cit:disk}
Micha{\l} Tyszkiewicz, Pascal Fua, and Eduard Trulls.
\newblock Disk: Learning local features with policy gradient.
\newblock \emph{NeurIPS}, 33:\penalty0 14254--14265, 2020.

\bibitem[Vaswani et~al.(2017)Vaswani, Shazeer, Parmar, Uszkoreit, Jones, Gomez,
  Kaiser, and Polosukhin]{cit:transformer}
Ashish Vaswani, Noam Shazeer, Niki Parmar, Jakob Uszkoreit, Llion Jones,
  Aidan~N Gomez, {\L}ukasz Kaiser, and Illia Polosukhin.
\newblock Attention is all you need.
\newblock \emph{NeurIPS}, 30, 2017.

\bibitem[Wu(2013)]{cit:visualsfm}
Changchang Wu.
\newblock Towards linear-time incremental structure from motion.
\newblock In \emph{3DV}, pages 127--134, 2013.

\bibitem[Zhang et~al.(2018)Zhang, Zhou, Lin, and Sun]{cit:shufflenet}
Xiangyu Zhang, Xinyu Zhou, Mengxiao Lin, and Jian Sun.
\newblock Shufflenet: An extremely efficient convolutional neural network for
  mobile devices.
\newblock In \emph{CVPR}, pages 6848--6856, 2018.

\bibitem[Zhao et~al.(2022)Zhao, Wu, Miao, Chen, Chen, and Li]{cit:alike}
Xiaoming Zhao, Xingming Wu, Jinyu Miao, Weihai Chen, Peter~CY Chen, and
  Zhengguo Li.
\newblock Alike: Accurate and lightweight keypoint detection and descriptor
  extraction.
\newblock \emph{IEEE TMM}, 2022.

\bibitem[Zhou et~al.(2021)Zhou, Sattler, and Leal-Taixe]{cit:patch2pix}
Qunjie Zhou, Torsten Sattler, and Laura Leal-Taixe.
\newblock Patch2pix: Epipolar-guided pixel-level correspondences.
\newblock In \emph{CVPR}, pages 4669--4678, 2021.

\end{thebibliography}
}

\newpage

\appendix

\twocolumn[{%
 \centering
 \LARGE [Supplementary Material] \\ XFeat: Accelerated Features for Lightweight Image Matching\\[1.5em]
}]

In this supplementary material accompanying the main paper, we present a more detailed overview of the architecture of our proposed CNN backbone and the practices employed in the training process. Moreover, we provide an expanded set of qualitative results and extended discussion, providing additional contextualization with the current state-of-the-art methods. Code and weights are available at \href{https://verlab.dcc.ufmg.br/descriptors/xfeat_cvpr24}{verlab.dcc.ufmg.br/descriptors/xfeat\_cvpr24}.

\section{Backbone details}\label{sec:backbone}

To maintain the backbone's structural simplicity, we employ a primary unit termed the basic layer. This unit is structured with a 2D convolution with square kernel sizes $k = 1$ or $k = 3$, complemented by ReLU activation and Batch Normalization. A stride of $2$ in the convolution is applied for halving the spatial resolution as needed. 
\begin{figure}[!b]
	\centering
	\includegraphics[width=0.999\columnwidth]{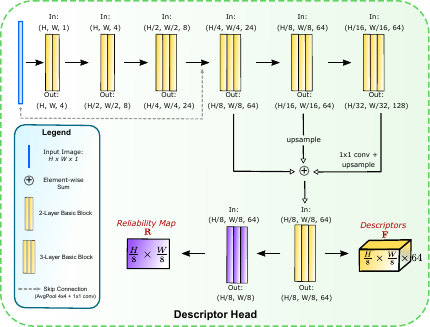}

	\caption{{\bf {Detailed descriptor backbone.}} 
     Our backbone is comprised of $23$ convolutional layers, following the downsampling strategy described in Sec. 3.1 of the main paper. Our network is deeper compared to ALIKE~\cite{cit:alike} and SuperPoint~\cite{cit:superpoint} backbones in terms of layers, but due to the efficient downsampling strategy adopted, our network's inference is much faster.
    }
	\label{fig:backbone}
\end{figure}
The network's architecture is modular, comprising several basic layers as a basic block, as depicted in~\cref{fig:backbone}. Each block consists of two or three basic layers. The backbone of our network comprises six of these basic blocks, designed to halve the spatial resolution in each step while progressively augmenting the depth using the approach detailed in Sec. 3.1 of the main paper. The first basic layer on each block performs the spatial downsampling. Two additional basic blocks, in the end, are employed to perform the fusion of multi-resolution features and reliability map prediction, respectively. Preliminary experiments revealed that adding a single skip connection to the model as shown in~\cref{fig:backbone} slightly increased performance, which has led to its incorporation in the final backbone design.


\begin{figure*}[tb!]
	\centering
	\includegraphics[width=0.8\textwidth]{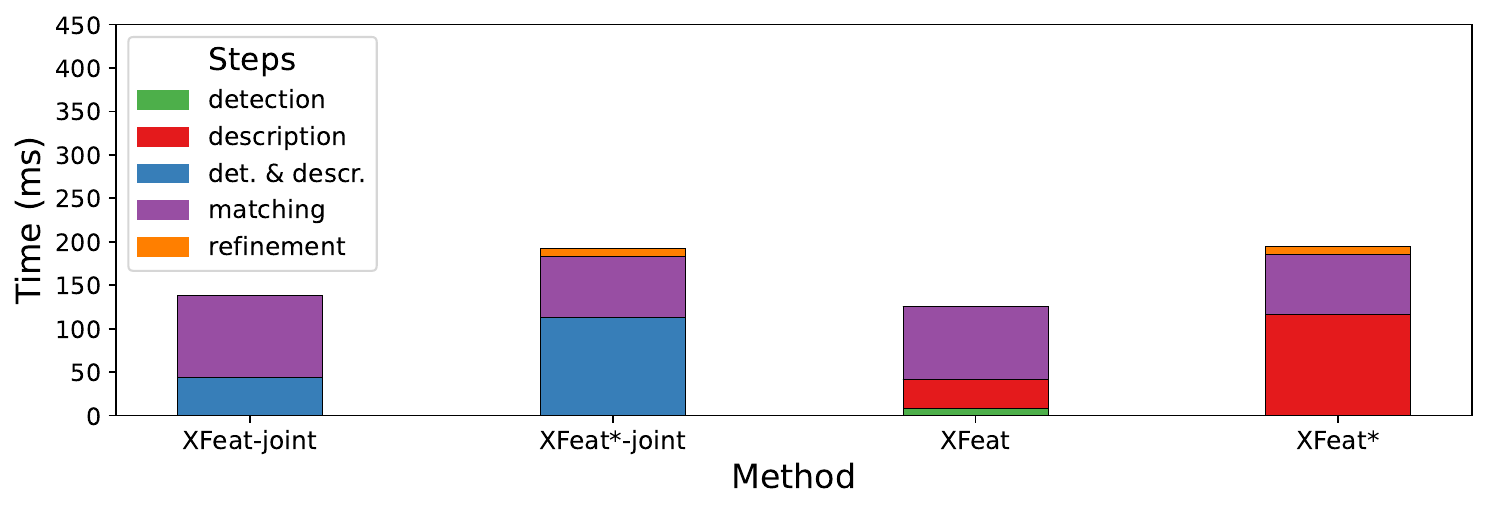}
	\vspace*{-0.4cm}
	\caption{{\bf{Detailed timing analysis on i7-6700K CPU.}} Required time by each step of our ablated methods. 
	}
	\label{fig:runtimes}
\end{figure*}

\section{ Training description }
 We trained the network on a mix of Megadepth~\cite{cit:megadepth} scenes using the training split provided by~\cite{cit:loftr} and synthetically warped pairs using raw images (without labels) from COCO~\cite{cit:coco} in the proportion of $6:4$ respectively. All image pairs were resized to $(W=800, H=600)$, and ground-truth correspondences were scaled accordingly. 
 Our ablations show that hybrid training significantly improves generalization for small CNNs, as observed in high-capacity models~\cite{cit:lightglue}. 
 The network was trained on batches of $10$ image pairs using the Adam optimizer~\cite{cit:adam} with an initial learning rate of $3\times 10^{-4}$, applying an exponential decay of $0.5$ at every $30{,}000$ gradient updates. Convergence is attained after $160{,}000$ iterations, within $36$ hours on a single NVIDIA RTX 4090 GPU, consuming $6.5$ GB of VRAM in total, considering both training and synthetic warps done on the fly on GPU. Disk I/O is the predominant speed bottleneck due to the overhead of loading images and depth maps from the Megadepth dataset in their original resolution, which can be easily solved with a more careful data preparation scheme. The low memory usage of our method enables training on entry-level hardware, facilitating the fine-tuning or full training of our network for specific tasks and scene types.


\section{Detailed timing analysis}

This section reports a detailed timing analysis of our proposed solutions in sparse and semi-dense matching settings.
Regarding XFeat$^*$'s match refinement step, we show in \cref{fig:runtimes} that the match refinement cost is negligible. More notably, even with the refinement step included, XFeat$^*$ achieves a similar matching time compared to XFeat with the same number of keypoints because refinement is performed after the nearest neighbor search. Additionally, we present the extraction running times for the most efficient methods available on an \textbf{Orange Pi Zero 3} equipped with a Cortex-A53 ARM processor. This device stands out as one of the \textbf{smallest and most affordable consumer-grade embedded computers} ($\$28$). Considering its limited processing power, we adjusted the input resolution to $480 \times 360$ for all methods and used their standard PyTorch implementation without any deployment optimization. Our findings show that XFeat operates at an average of $1.8$ FPS, SuperPoint at $0.16$ FPS, and ALIKE at $0.58$ FPS, respectively. This experiment shows that XFeat is the only learned method capable of running over one FPS on a highly constrained embedded device that is not optimized for neural network inference.

\section{Megadepth-1500 qualitative results}
\begin{figure*}[tb!]
	\centering
	\includegraphics[width=0.999\textwidth]{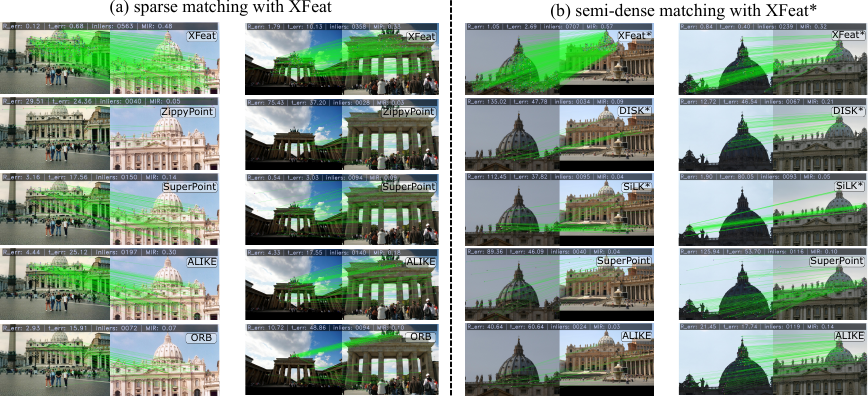}

	\caption{{\bf {Additional qualitative results on Megadepth-1500~\cite{cit:megadepth, cit:loftr} landmark dataset.}} 
  XFeat and XFeat$^{*}$ are robust in demanding scenarios with significant viewpoint and illumination variations, outperforming even the more computationally intensive DISK model in semi-dense matching with $10{,}000$ local features at a striking $16\times$ speedup. In a sparse setting with $4{,}096$ keypoints, our method, which is many times faster than ALIKE ($5\times$) and SuperPoint ($9\times$), demonstrates more robustness to wide baseline transformations due to the effective re-formulation of XFeat's backbone CNN. 
    }
	\label{fig:quali}
\end{figure*}

\cref{fig:quali} shows more qualitative results of our two proposed approaches compared to the baseline methods used in the main paper. For more challenging cases such as strong viewpoint and illumination changes, XFeat and XFeat$^{*}$ exhibit exceptional robustness even compared to DISK~\cite{cit:disk} -- the largest CNN architecture regarding floating point operations. We hypothesize that this robustness is attributed to our network's large receptive field and depth compared to shallower models such as SuperPoint, ALIKE, and SiLK~\cite{cit:silk}, demonstrating the effectiveness of our featherweight backbone in the compute-accuracy trade-off.

\section{ScanNet-1500 extended discussion} \label{sec:scannet}
Recalling the results obtained in Tab. 2 of the main paper, XFeat and XFeat$^{*}$ surpass both fast and standard local feature extractors in pose accuracy while being significantly faster for indoor relative pose estimation. DISK and ALIKE, which were trained in the same Megadepth scenes as XFeat, display signs of overfitting in landmark imagery: they perform exceptionally well in strict thresholds (AUC@5$^{\circ}$) on Megadepth-1500 test set, but their relative performance are similar or worse in tasks such as homography estimation and visual localization compared to XFeat and SuperPoint, as one can observe in Tab. 3 and Tab. 4 of the main paper.

We conjecture that XFeat produces less biased local descriptors due to our hybrid training with synthetic warps on COCO. SuperPoint also demonstrate increased generalization accross different downstream tasks and datasets due to its inherent self-supervised training strategy on synthetic warps. Hybrid training can encourage local feature representations to focus less on distinctive textures often present in landmark outdoor imagery that could bias the CNN training. In addition, the large receptive field of our network, as well as its increased layer depth compared to the other approaches, helps XFeat in indoor imagery (which often lacks distinctiveness at the local level), resulting in more consistent matches compared to DISK and ALIKE in ScanNet-1500, even though XFeat and the competitors were not trained on ScanNet data.

\begin{figure*}[tb!]
	\centering
	\includegraphics[width=0.99\textwidth]{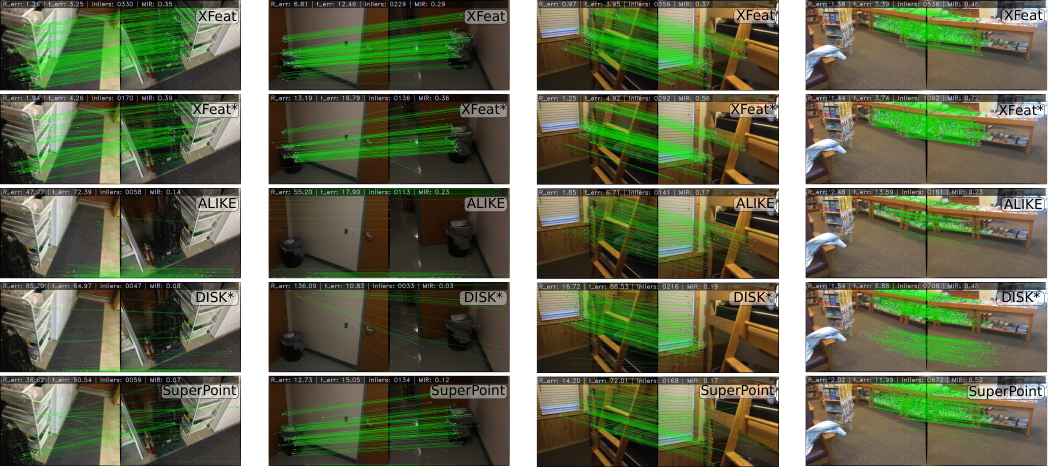}

	\caption{{\bf {Additional qualitative results on ScanNet-1500~\cite{cit:scannet, cit:loftr} indoor dataset.}} 
   Our proposed approaches consistently outperform state-of-the-art methods such as DISK and ALIKE in indoor imagery, both in terms of camera pose and inlier ratio. Notice that SuperPoint also often outperforms DISK and ALIKE. \cref{sec:scannet} provides a detailed discussion on the reasons behind our method's superiority.
    }
	\label{fig:quali_scannet}
\end{figure*}

\section{Comparison with learned matchers}

\begin{table*}[tb!]
\centering
\caption{{\bf{Matchers comparison on Megadepth-1500.}} Inference speed in pairs per second (PPS) @ $1{,}200$ px. (i7-6700K CPU).}
\label{tab:learned_matcher_eval}
\resizebox{0.85\textwidth}{!}{%
\begin{tabular}{@{}lcccc|ccrrr@{}}
\toprule
\textbf{Method} & \textbf{Type} &\textbf{AUC@$5^\circ$} & \textbf{@$10^\circ$} & \textbf{@$20^\circ$} & \textbf{Acc@$10^\circ$} & \textbf{MIR} & \textbf{\#inliers} & \textbf{PPS} \\
\midrule

\textcolor{black}{LoFTR}& \textcolor{black}{learned matcher}        & \textcolor{black}{68.3} & \textcolor{black}{80.0} & \textcolor{black}{88.0} & \textcolor{black}{93.9} & \textcolor{black}{0.93} & \textcolor{black}{3009} & \textcolor{black}{$0.06$}  \\

\textcolor{black}{LightGlue} & \textcolor{black}{learned matcher}    & \textcolor{black}{61.4} & \textcolor{black}{75.0} & \textcolor{black}{84.8} & \textcolor{black}{91.8} & \textcolor{black}{0.92} & \textcolor{black}{475} &  \textcolor{black}{0.31} \\

\hdashline \noalign{\vskip 0.5ex}
Patch2Pix & coarse-fine   & \underline{47.8} & \underline{61.0} & \underline{71.0} & \underline{77.8} & \underline{0.59} & \underline{536} &  \underline{$0.05$}  \\
XFeat$^{*}$ & coarse-fine   & \bf{50.2} & \bf{65.4} & \bf{77.1} & \bf{85.1} & \bf{0.74} & \bf{1885} & \bf{1.33}  \\

\bottomrule
\end{tabular}
}
\end{table*}

 Since XFeat$^{*}$ uses paired inputs when performing the refinement step, we provide additional comparisons of XFeat$^{*}$ (semi-dense matching) with popular learned matchers such as LoFTR~\cite{cit:loftr} and LightGlue~\cite{cit:lightglue}, and coarse-to-fine strategies as Patch2Pix~\cite{cit:patch2pix}, to elucidate the key differences. The results for these new approaches are shown in \cref{tab:learned_matcher_eval}. Although XFeat$^*$ needs paired inputs for refinement, it fundamentally differs in its methodology from learned matchers, being only comparable to Patch2Pix, as we rely on traditional nearest neighbor search for matching, followed by a lightweight refinement of matches, incurring a negligible computational load (see \cref{fig:runtimes}). The requirement for paired inputs does not change the usual pipeline for SfM and visual localization tasks because XFeat$^*$'s features can be stored for each image independently, as usually done for sparse settings. For instance, high-resolution feature maps are not required, unlike LoFTR, to produce refined matches.
 
 Our techniques are, in fact, complementary to learned matchers; for example, LightGlue can be trained using both XFeat and XFeat$^{*}$ features. Learned matchers are more data hungry and much more expensive to train, \textit{e.g.}, LoFTR uses $64$ GPUs for $24$ hours to be trained. XFeat$^{*}$, for its turn, can be trained on a single 8 GB GPU. Furthermore, XFeat$^*$ offers up to $22 \times$ speedup over existing semi-dense solutions as shown in \cref{tab:learned_matcher_eval} and surpasses coarse-to-fine approaches such as Patch2Pix in accuracy, while being faster and delivering many more matches than sparse learned matchers as LightGlue. Naturally, XFeat, as a local descriptor, offers limited robustness to aggressive viewpoint changes and highly ambiguous image pairs compared to transformer-based feature matchers. Coupling a lightweight transformer such as LightGlue or LoFTR's linear transformer with XFeat's local features can open new directions in scalable, high-performance image matching tasks, facilitating advancements in both efficiency and accuracy that are pivotal for pushing the boundaries in visual navigation, augmented reality, and real-time visual SLAM.


\end{document}